\documentclass[10pt, conference, compsocconf, letterpaper]{IEEEtran}
\usepackage[T1]{fontenc}
\usepackage{graphicx,tabularx,siunitx,oplotsymbl,xcolor,color,soul,booktabs,balance,adjustbox,hyperref,lipsum,multirow,bbold,tikz,amsmath,amsfonts,mathtools,microtype,bm,array,threeparttable,float,subfig}
\usepackage{caption}
\usepackage{nth}
\usepackage{multirow}
\usepackage{makecell}
\usepackage{enumitem}
\usepackage{cite}
\usepackage{amssymb,amsfonts}
\usepackage{textcomp}
\usepackage[algo2e,linesnumbered,ruled]{algorithm2e}

\usepackage[capitalise,noabbrev]{cleveref}
\usepackage{url}

\IEEEoverridecommandlockouts
\def\BibTeX{{\rm B\kern-.05em{\sc i\kern-.025em b}\kern-.08em
    T\kern-.1667em\lower.7ex\hbox{E}\kern-.125emX}}

\newcommand{\systemName}{UR2M}
\newcommand{\emphasize}[1]{\textbf{\textit{#1}}}

\SetCommentSty{mycommfont}
\SetKwInput{KwInput}{Input}                % Set the Input
\SetKwInput{KwOutput}{Output}              % set the Output
\SetKwInput{KwRequire}{Require}              % set the Require

\begin{document}\sloppy

\title{\systemName: Uncertainty and Resource-Aware Event
    Detection on Microcontrollers
    % \vspace{-1.5em}
    }
% \vspace{-1em}
\author{\IEEEauthorblockN{Hong Jia\IEEEauthorrefmark{1}, Young D. Kwon\IEEEauthorrefmark{1}, Dong Ma\IEEEauthorrefmark{2}, Nhat Pham\IEEEauthorrefmark{3}, Lorena Qendro\IEEEauthorrefmark{4}, Tam Vu\IEEEauthorrefmark{5} and Cecilia Mascolo\IEEEauthorrefmark{1}} \IEEEauthorblockA{\IEEEauthorrefmark{1}University of Cambridge, Cambridge, UK \IEEEauthorrefmark{2}Singapore Management University, Singapore\\
\IEEEauthorrefmark{3}Cardiff University, Cardiff, UK
\IEEEauthorrefmark{4}Nokia Bell Labs, Cambridge, UK
\IEEEauthorrefmark{5}University of Colorado Boulder, Colorado, US
\\
\{hj359, ydk21\}@cam.ac.uk, dongma@smu.edu.sg, phamn@cardiff.ac.uk, \\ lorena.qendro@nokia-bell-labs.com, tam.vu@colorado.edu, cm542@cam.ac.uk}
}
\vspace{-1em}

\maketitle
\thispagestyle{plain}
\pagestyle{plain}
\pagenumbering{gobble}
\begin{abstract}
    Traditional machine learning techniques are prone to generating inaccurate predictions when confronted with shifts in the distribution of data between the training and testing phases. This vulnerability can lead to severe consequences, especially in applications such as mobile healthcare. Uncertainty estimation has the potential to mitigate this issue by assessing the reliability of a model's output. However, existing uncertainty estimation techniques often require substantial computational resources and memory, making them impractical for implementation on microcontrollers (MCUs). This limitation hinders the feasibility of many important on-device wearable event detection (WED) applications, such as heart attack detection.
    
    In this paper, we present \textbf{\systemName{}}, a novel \textbf{Uncertainty} and \textbf{Resource-aware} event detection framework for \textbf{MCUs}. Specifically, we (i) develop an uncertainty-aware WED based on evidential theory for accurate event detection and reliable uncertainty estimation; (ii) introduce a cascade ML framework to achieve efficient model inference via early exits, by sharing shallower model layers among different event models; (iii) optimize the deployment of the model and MCU library for system efficiency. We conducted extensive experiments and compared \systemName{} to traditional uncertainty baselines using three wearable datasets. Our results demonstrate that \systemName{} achieves up to 864\% faster inference speed, 857\% energy-saving for uncertainty estimation, 55\% memory saving on two popular MCUs, and a 22\% improvement in uncertainty quantification performance. 
    \systemName{} can be deployed on a wide range of MCUs, significantly expanding real-time and reliable WED applications.
\end{abstract}

\begin{IEEEkeywords}
    Uncertainty, Event Detection, Efficiency, Microcontrollers
\end{IEEEkeywords}

% \vspace{-1em}
\section{Introduction}
\label{introduction}

\begin{figure}[t]
  % \vspace{-1em}
  \centering
  \includegraphics[trim={0cm 0.2cm 0cm 0cm},clip,width=\columnwidth]{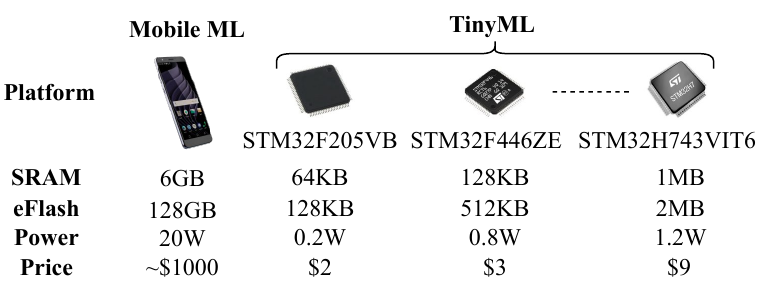}
  \caption{Memory and power comparison between a typical mobile phone and microcontrollers.}
  \vspace{-1em}
  \label{fig:mcu}
\end{figure}

With advancements in pervasive, low-power, and embedded sensors, a range of human physiological signals can be collected and continuously analyzed. Empowered by machine learning (ML), especially deep learning (DL), these sensors provide great opportunities for a plethora of wearable event detection (WED) applications, such as the detection of stress levels~\cite{alavi2022real}, blood pressure~\cite{10.1145/3132024}, or respiratory illnesses~\cite{10.1145/3274783.3275159}. Recently, deploying ML models directly on microcontrollers (MCUs) has attracted tremendous attention due to their potential to improve user privacy and computational latency in WED, especially under unstable network conditions~\cite{lin2020mcunet}. However, as shown in Figure~\ref{fig:mcu}, designing and deploying efficient WED models on MCUs is challenging due to their limited memory space and battery life, especially in comparison to mobile phones~\cite{lin2020mcunet}.

Furthermore, many existing WED models prioritize enhancing classification accuracy while overlooking the importance of prediction reliability~\cite{ovadia2019can}, which is crucial in fields like health. Reliability is quantified as \textit{uncertainty}, indicating the trustworthiness of the classification results~\cite{carneiro2020deep}. Factors such as hardware differences, environmental variations, data collection methods, and sensor degradation can lead to distribution shifts between training and testing data (data uncertainty) or unseen data (model uncertainty~\cite{lakshminarayanan2017simple}), reducing the reliability of WED models.

Several methods for quantifying uncertainty have been investigated. Bayesian Neural Networks (BNNs), a prominent approach for uncertainty estimation, quantify uncertainty by estimating posteriors over model weights~\cite{10.1145/3560905.3568417}. However, BNNs entail substantial computational expenses~\cite{qendro2021benefit}. Although approximation techniques such as Monte Carlo dropout (MCDP)~\cite{gal2016dropout} and deep ensembles~\cite{lakshminarayanan2017simple} have been proposed, these methods still require ensembling multiple models and various inference steps, which introduce intensive computational and memory demands, as well as increased latency. Recent research has also introduced deterministic models that require only one forward pass, making them more efficient but at the cost of lower accuracy~\cite{liu2020simple}. As a result, integrating reliable uncertainty could pose additional complexities in the design and deployment of trustworthy WED models on MCUs. 

Lastly, existing works demonstrate inefficiency in supporting multi-event detection on MCUs, as they typically employ individual models for each event to ensure reusability across different applications or use cases and to optimize efficiency for each model~\cite{pham2022pros}. However, wearable devices often require the simultaneous detection of multiple events. For instance, a single electroencephalography (EEG) input might be utilized to concurrently detect the brain's alpha wave (event 1) for a guided-meditation application, and beta wave (event 2) for a focus monitoring application. Additionally, executing multiple inferences (encompassing both prediction and uncertainty estimation) for varied events can be resource-intensive, potentially rendering WED deployment on MCUs impracticable due to memory constraints. 

To address the aforementioned challenges, we propose an efficient uncertainty estimation approach based on evidential deep learning (EDL) and cascade learning. Specifically, (i) EDL is designed to predict a distribution, parameterized by a vector, instead of providing a point prediction through a single DL model, which allows for the direct prediction of event detection and its associated uncertainty via a single inference. (ii) For each event (intra-event), we consider three models of varied depths (i.e., shallow, medium, and deep); herein, deeper models are stacked upon shallower ones, meaning the lower layers are shared. A classifier layer (termed a ``head'') is appended to each model. This design adheres to the observation that some testing samples, particularly those near the center of the training sample distribution, do not require a full pass through the deep model to ensure a reliable prediction~\cite{teerapittayanon2016branchynet}. Consequently, early exits can be employed to enhance computational cost-effectiveness and inference speed, with uncertainty chosen as the criterion for an early exit to ensure the reliability of the prediction. (iii) For multiple events (inter-event) using the same input, we propose the sharing of all layers for feature extraction and the training of individual classification layers (referred to as ``multi-heads''). As a result, our framework can be effortlessly scaled to multiple events with minimal memory overhead, since only the heads need to be added. Additionally, reusing shared layers for different events reduces computation time and cost.

We further apply three techniques to improve the efficiency of our approach during implementation. First, we implement an architecture search to find the optimal model structure automatically (e.g., number of model layers and size of channels) for specific WED tasks based on recent success models designed for MCUs~\cite{banbury2021micronets}. Second, we conduct scalar quantization of the model weights into 8-bit integers to decrease the model size and further save memory. Third, to reduce the memory consumption of the deep learning library, we remove unnecessary components that are not utilized in our models. Finally, we conduct comprehensive experiments with two MCU platforms to demonstrate the effectiveness of the proposed approach.

To summarize, we make the following contributions:
\begin{itemize}
  \item We propose a cascade model architecture with intra-event and inter-event layer sharing to enable efficient multi-event detection. We also conduct efficient architecture search, model compression, and library optimization to improve system efficiency (\S\ref{sec:architecture}-\S\ref{sec:implementation}).

  \item We propose a novel uncertainty-aware learning paradigm based on evidential theory for efficient and reliable WED uncertainty estimation on MCUs (\S\ref{sec:uncertainty quantification}).

  \item We conduct extensive experiments on three popular wearable datasets and implement our framework on two off-the-shelf MCUs, including STM32F446ZE and STM32H747F7, with limited SRAM memory (128KB and 512KB, respectively). Our evaluation shows that the proposed framework performs up to 864\% better inference speed and 857\% energy saving compared to uncertainty baselines.
        The approach also saves 55\% of memory compared with existing uncertainty estimation baselines (\S\ref{sec:evaluation}-\S\ref{sec:results}), enabling the deployment of WED models on MCUs with limited memory (e.g., STM32F205VB with 64KB SRAM).
\end{itemize}
% \vspace{-1em}
\section{Related Works}
\label{sec:relatedwork}
This section briefly discusses the literature on machine learning on MCUs, event detection on resource-constrained devices, and efficient methods for uncertainty estimation.

\begin{figure*}[t]
  \centering
  \subfloat{
    \includegraphics[trim={0cm 0cm 0cm 0cm},clip, height=5.8cm]{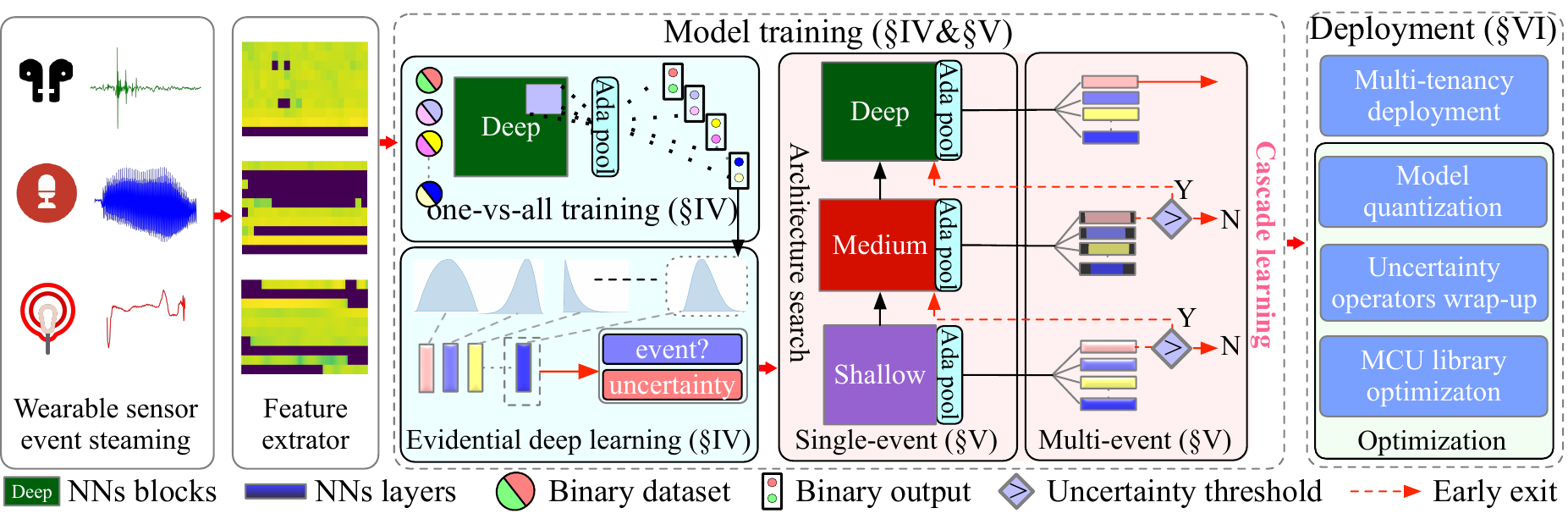}}
    \label{fig:frameworkstage1}
  
  \caption{System overview.}
  \label{fig:systemoverview_orig}
  \vspace{-1.5em}
\end{figure*}

\emphasize{Tiny machine learning on MCUs.}
Tiny Machine Learning~\cite{banbury2021micronets} (TinyML) aims to execute deep learning models locally on extremely resource-constrained devices such as MCUs. Recent studies have concentrated on optimizing network architectures considering constraints such as limited memory, energy, FLOPs~\cite{lin2020mcunet}, and processor speed~\cite{liberis_unas_2021}. However, these approaches focus solely on classification accuracy, treating them as single-point predictions without considering uncertainty estimation. In contrast, we further include uncertainty estimation of the desired predictions to enable a more reliable~WED.

\emphasize{Event detection on resource-constrained devices.}
Recent years have seen a surge in research focused on event detection using wearables, exploring various sensing modalities including image~\cite{ghodrati2021frameexit}, audio~\cite{bondareva2021earables}, electrocardiogram (ECG)~\cite{10.1145/3485730.3492883}, and others. However, most existing WED approaches only utilize wearables for data collection, offloading processing tasks like pre-processing, feature extraction, and ML modelling to cloud-based GPUs (through WiFi)~\cite{10.1145/3274783.3275159,lee2019occlumency}, desktop GPUs~\cite{10.1145/3463511}, mobile devices~\cite{alavi2022real} or IoT devices~\cite{huang2020clio}. This category of approaches can lead to high latency during signal transmission or raise privacy concerns. To address these challenges, our focus is on comprehensive WED for on-MCU computation, developing efficient and lightweight ML models suitable for limited-resource environments.

\emphasize{Efficient uncertainty estimation.}
Some effort has been devoted to achieving efficient uncertainty estimation, such as regulating the neural network weights to simulate BNNs~\cite{sensoy2018evidential}. Another stream of studies focuses on expensive and not deployable operations on MCUs like flow~\cite{malinin2018predictive}, spectral normalization~\cite{mukhoti2021deep}, and stochastic Convolutional layers~\cite{qendro2021benefit}. Despite their success in improving computation efficiency, their accuracy still either performs four times worse than the state-of-the-art (SOTA) method of deep ensembles~\cite{lakshminarayanan2017simple} or require customized operators and libraries that are currently unavailable on MCUs. As an alternative to using ensembles, knowledge distillation~\cite{malinin2019ensemble} has been proposed as a means of training a single model. However, knowledge distillation typically requires out-of-distribution (OOD) data, which is often difficult to obtain for real-world applications. Compared to existing work, our study is the first to propose an efficient model for uncertainty quantification on MCUs.

\vspace{-0.5em}
\section{\systemName{} System overview}
\label{fig:overview}
\systemName{} includes two stages: \textbf{model training} ($\S$\ref{sec:uncertainty quantification}-$\S$\ref{sec:architecture}) and \textbf{deployment} ($\S$\ref{sec:implementation}) as shown in Figure~\ref{fig:systemoverview_orig}. During the \textbf{training stage}, there are three objectives: (1) EDL for efficient uncertainty quantification, (2) Cascade ML learning which includes single-event (intra-event) detection via early exits, and multi-event (inter-event) detection via feature sharing and multi-heads. During the \textbf{deployment stage}, we first carry out (1) multi-tenancy deployment~\cite{robert_tflm_mlsys2021}, allowing multiple ML models (referred to as ``tenants'') to efficiently and dynamically share the same memory space among intra-event models. We then further focus on (2) optimizing the model and the MCU library.

In detail, wearable sensors first capture event streaming signals. Features are then extracted for different signals, such as Mel-frequency cepstral coefficients (MFCC) for the audio signals. Following this, evidential modeling via EDL and one-vs-all training \textbf{(\S\ref{sec:uncertainty quantification})} are applied to obtain reliable WED predictions and estimate uncertainty. Within the EDL framework, we specifically designed a cascade learning architecture \textbf{(\S\ref{cascade learning})} for single-event detection, which divides the network layers into shallow, medium, and deep levels to enable intra-event sharing (sharing shallower layers and inferring with early exits within an event model) and process samples at different levels of recognition difficulty. Further, we propose inter-event sharing (sharing entire layers for feature extraction) for multi-event detection. In addition to the modeling, we further carry out efficiency improvements \textbf{(\S\ref{sec:implementation})} via model architecture search (during model training), quantization, uncertainty operator wrap-up, and MCU library optimizations.
%!TEX root = main.tex
% \vspace{-0.5em}
\section{Efficient Uncertainty Quantification}\label{sec:uncertainty quantification}
In this Section, we propose a highly efficient EDL model tailored for event detection on MCUs. This model is optimized to adhere to the constraints of MCUs, employing distributions to achieve accurate uncertainty quantification in real-time scenarios through a single forward pass.

\subsection{Evidential Deep Learning}\label{sec:edl}

For a given input $x^{i}$, EDL generates a Dirichlet distribution $Dir(\boldsymbol{\alpha}^{i})$, 
 where $\boldsymbol{\alpha}^{i}=[\alpha_1^{i}$,$\alpha_2^{i}$,...,$\alpha_C^{i}]$ denotes the concentration parameters of the distribution (dense distribution means high evidence and low uncertainty)~\cite{sensoy2018evidential}. Being a conjugate prior to the categorical distribution, the Dirichlet distribution enables EDL to determine the belief mass $\boldsymbol{b}^{i} =[b_1^{i}$,$b_2^{i}$,...,$b_C^{i}]$ correlating directly with uncertainty. A higher belief mass indicates a higher confidence in the prediction, whereas a lower belief mass suggests the presence of uncertainty. Formally,
\begin{equation}
    \label{eq:belief_mass}
    \boldsymbol{b}^{i}=(\boldsymbol{\alpha}^{i}-1)/S^{i},
\end{equation}
where $S^{i} = \sum_{c=1}^C\alpha_{c}^{i}$ is the Dirichlet strength. From  $\boldsymbol{\alpha}^{i}$ and $\boldsymbol{b}^{i}$, we can further infer the categorical prediction $\hat{y}^{i}$ and the associated uncertainty $u^{i}$ as:
\vspace{-1em}
\begin{equation}
    \label{eq:dirch_uncertainty}
    \hat{y}^{i}=\arg \max _c [{\alpha}^{i} /S^{i}], \quad u^{i} = 1-\sum_{c=1}^{C} {b}_{c}^{i}
\end{equation}
\vspace{-1em}

Before the training process, acknowledging our initial state of complete uncertainty about the outputs (i.e., uncertainty $u^i$ is set to 1), we initialize $\boldsymbol{\alpha}^i$ with $[1,1,1]$, corresponding to $\boldsymbol{b}^i = [0,0,0]$ according to Eq.~\ref{eq:belief_mass} and Eq.~\ref{eq:dirch_uncertainty}. To refine the model, we employ a loss function defined as:

\vspace{-1.2em}
\begin{equation}
    \label{eq:loss}
    \min _\theta \mathcal{L}=\frac{1}{N} \sum_i^N C E(\alpha_c^i / S^i, y^{i})-\lambda \cdot H( Dir(\boldsymbol{\alpha}^i))
\end{equation}
\vspace{-1.2em}

where $CE$ denotes the cross-entropy loss, and $H$ represents the entropy of a Dirichlet distribution parameterized by $\boldsymbol{\alpha}^i$. The first term of the loss function aims to maximize classification accuracy, while the second term controls the output distribution to avoid overconfidence. The hyperparameter $\lambda$ plays a crucial role in balancing these two terms.

Finally, this procedure will lead to a predicted $\boldsymbol{\alpha}^i$ for each sample which is used to infer the categorical outcome and the associated uncertainty (e.g., $u = 1-\sum b^i$).

\subsection{Efficient Evidential Modeling for Event Detection on MCUs}
\label{subsec:evidentialmodeling}

Implementing the EDL discussed in \S\ref{sec:edl} for WED requires deploying multiple models and performing a series of inferences to detect various events, which significantly challenges the limited computational resources of MCUs. To mitigate this, we propose an efficient EDL modeling for WED, along with related training and optimization techniques designed to infer multiple events concurrently.

\textbf{\textit{Efficient EDL Modeling for WED}.} WED is designed to identify an event signal coming from a wearable device. In ML/DL, this objective is defined as a binary classification task over a given duration/period of sensor data. For each binary classifier that detects classes of the event $c$, the outputs of EDL include the binomial belief mass, which can be used to infer the uncertainty of the WED prediction, i.e., how confident it is to be classified as positive (i.e., an event happening) or negative (i.e., an event not happening).

Given the binary nature of our EDL framework (positive vs negative), we adopt a Beta distribution (a special case of the Dirichlet distribution) to model the event probability. Specifically, a Beta distribution is characterized by two parameters $\alpha_c^i$ and $\beta_c^i$ such that 

\vspace{-1em}
\begin{equation}
    \begin{split}
        \mathrm{P}({p_c^{i}}\mid{x}^i ; {\theta_c})&=\operatorname{Beta}(p_c^{i} \mid \alpha_c^{i}, \beta_c^{i}) \\
        &=\frac{1}{B(\alpha_c^{i}, \beta_c^{i})} p^{\alpha_c^{i}-1}(1-p)^{\beta_c^{i}-1},
    \end{split}
\end{equation}
% \vspace{-1em}

where $\mathrm{P}\left({p_c^{i}}\mid{x}^i ; {\theta_c}\right)$ denotes the probability distribution of the event given the sensor sample ${x}^i$, with both 
$\alpha_c^{i}$, and $\beta_c^{i}$ being greater than zero. $B(\alpha_c^{i}, \beta_c^{i})=\Gamma(\alpha_c^{i}) \Gamma(\beta_c^{i}) / \Gamma(\alpha_c^{i}+\beta_c^{i})$ is the Beta function, $\Gamma(\cdot)$ is the gamma function, and $p_c^{i}\neq0$. 
Applying the mapping rule in Eq.~\ref{eq:dirch_uncertainty}, the belief mass and uncertainty $u$ for each sample $i$ are derived via a NN:

\vspace{-1em}
\begin{equation}
    \label{beta_acc}
    {b}_{1}^i=\frac{\alpha_c^{i}-1}{\alpha_c^{i}+\beta_c^{i}}, \hspace{0.5em} {b}_{2}^i=\frac{\beta_c^{i}-1}{\alpha_c^{i}+\beta_c^{i}}
\end{equation}
% \vspace{-1em}
\begin{equation}
    \label{beta_uncertainty}
    {u}^i={2}/{(\alpha_c^{i}+\beta_c^{i})}
\end{equation}
\vspace{-1em}

where ${b}_{1}$ represents the belief mass of a positive prediction while ${b}_{2}$ denotes that of a negative prediction.

\textbf{\textit{One-versus-all classifiers}}. To obtain the parameters of $\alpha_c^i$ and $\beta_c^i$ in EDL across multiple events, we adopt the one-versus-all (OVA) classifier, where each classifier distinguishes a specific event from all others, leading to $C$ binary classifiers (i.e., heads). Specifically, in multi-event WED, we split the entire training dataset into $C$ independent datasets with binary labels (i.e., event $c$ vs. non-event $c$ for $c \in [1, C]$). For each event, we then develop a model to learn a set of mapping functions $h_c({x}^i;\theta_c)$, where ${x}^i$ represents the input signal, and $\theta_c$ are the model weights. The outputs of the mapping functions yield the parameters $\alpha_c^i$ and $\beta_c^i$ in the Beta distribution, computed as:

\vspace{-1em}
\begin{equation}
    \quad {\alpha_c^{i},\beta_c^{i}}={h}_{c}\left({x}^i ; {{\theta_c}}\right)
\end{equation}
\vspace{-1em}

From this, we can deduce binomial decisions, with $b_1^i$  denoting a positive prediction (i.e., event happening), and $b_2^i$ representing a negative prediction (i.e., event not happening). Subsequently, these mapping functions are optimized jointly through an OVA training~\cite{padhy2020revisiting}. With this joint training of a shared EDL model, there is no need to deploy separate models on MCUs, thereby significantly reducing memory costs.

In contrast to traditional softmax-based deep learning approaches, which force the Neural Networks (NNs) to predict a point estimation, we can replace the softmax layer of the neural network with a ReLU layer (or an exponential function but softplus is not available in the MCU library). This adjustment ensures that the outputs remain non-negative, aligning with the positive ${\alpha}^i$ and enabling the NNs to predict distributions for each event task.

\subsection{Uncertainty-aware training and optimization}
\label{subsec:evidentialtraining}

Focusing on the training and optimization of the EDL framework for the proposed multi-event WED, we draw inspiration from Eq.~\ref{eq:loss} and propose using the binary cross entropy and Beta loss for each binary classifier of event $c$ as:
% related loss function to capture To joint training the multiple events of on/off signal of each event of interest every $T$ timestamp on WED, we formalize an OVA training and optimization problem denoted as ${\operatorname{argmax}}(h_c(\boldsymbol{x}^i))$, where each binary classifier $h_c(x)$ can detect related event. 

\vspace{-1.5em}
\begin{equation}
    \min _\theta \mathcal{L}=\frac{1}{N} \sum_i^N BCE\left(\psi_c^i / S_c^i, y_c^i\right)-\lambda \cdot H\left(B\left(\psi_c^i\right)\right)
\end{equation}

where $\psi_c^i$ symbolizes the Beta distribution parameters $(\alpha_c^i, \beta_c^i)$, $\text{BCE}$ is the binary cross-entropy loss, $H$ represents the entropy of a Beta distribution $B$ parameterized by $\psi_c^i$ and $\lambda$ serves as a balancing weight between the cross-entropy loss and entropy of the Beta loss. For all $C$ events, we collectively optimize all binary classifiers~\cite{franchi2020one}, enabling the model to perform inference with just a single forward pass.

\section{Cascade learning}
\label{cascade learning}

This section discusses designing efficient neural networks for \systemName{}. We explore the benefits of the early-exit strategy and architecture search method for single-event sharing on MCUs, reducing computational and memory costs. We also examine multiple-event sharing and detail the training pipeline using cascade learning, with all search and training \textit{on the server}.
\label{sec:architecture}

\subsection{Single-event Sharing}
\label{subsec:neuralnetworkmotivation}
For many DL tasks, some input samples, referred to as ``easy'' samples, can be effectively classified using shallower layers of the representation. This indicates that these shallower representations can identify ``easy'' samples, thus avoiding extra computation, whereas more ``difficult'' samples require processing through deeper layers~\cite{dai2020epnet}. However, unlike edge GPUs, designing model sharing on MCUs is challenging given the limited computing power, memory, and library support.

\textbf{\textit{Using Early-exits to Share Shallower Layers.}}
We propose a nested architecture featuring three early exits (sub-networks), which include shallow, medium, and deep models designed for single-event (intra-event) sharing, as illustrated in Figure~\ref{fig:systemoverview_orig} for MCUs. Each sub-network is designed using identical blocks of neural network layers, inspired by efficient neural networks for edge devices~\cite{sandler2018mobilenetv2}. Existing early-exit methods usually rely on accuracy as a criterion to prune model branches. However, uncertainty can act as a crucial indicator for reliable prediction: we propose using uncertainty as a metric to determine whether to exit at each sub-network. As demonstrated in Figure~\ref{fig:systemoverview_orig}, uncertainty thresholds are applied at the output of both shallow and medium models to facilitate early exits for data with low uncertainty (i.e., reliable predictions), thereby saving on MCU overheads. 

\textbf{\textit{Uncertainty-aware Architecture Search}.} To find efficient neural networks that minimize MCU overhead, recent studies have shown that the number of operations (OPS) and channel sizes~\cite{banbury2021micronets} are two crucial factors. Considering this, we propose an effective yet straightforward architecture search method to identify optimal neural networks for the early-exit models (i.e., shallow, medium, and deep models) in single-event sharing.

\setlength{\textfloatsep}{0.1cm}
\begin{algorithm2e}[t]
    \caption{The Search and training of \systemName{}}
    \label{alg:search}
    \SetAlgoNoLine
    \DontPrintSemicolon
    \KwIn{Channel $L$, OPS size $O$, $\mathcal{D}^{TRAIN}$, $\mathcal{D}^{TEST}$}
    \KwOut{Event prediction $y$ and uncertainty $u$}
    \KwData{Training data $\mathcal{D}^{TRAIN}$}
    \tcc{search single-event model}
    \LinesNumbered
    best\_backbone, best\_score = False, 0\;
    \SetAlgoVlined
    \For{$i$ in L}{
        \SetAlgoVlined
        \For{$j$ in O}
        {
            \tcp{Train candidate NNs backbone ($\mathbf{b}_{ij}$)}
            NN$\leftarrow$$\mathbf{b}_{ij}(\mathbf{W}_{ij}, L_{i}, O_{j})$\;
                accuracy $\leftarrow$ NN($\mathcal{D}^{TRAIN}$)\;
                tradeoff $\leftarrow$ accuracy/OPS\;
            \If{tradeoff $<$ best\_score}
            {best\_NN, best\_score = NN, tradeoff}
            \Return{best\_NN}
        }
    }
    \tcc{train with cascade learning}
    \For{$l = 0, 1, 2$}{
        \tcp{take each output as next exit's input}
        $u$, output $\leftarrow$$(\mathbf{b}_{l}\;(\mathbf{W}_{l}),\mathcal{D}^{TRAIN}$)\;
        $\mathcal{D}^{TRAIN}$ $\leftarrow$ output\;
        \If{converge}{
            \Return{$\mathbf{W}_{l}$}
        }
    }
\end{algorithm2e}
\setlength{\floatsep}{0.1cm}

Specifically, we employ the Depthwise block as the OPS to control model depth, as it serves as an ideal proxy for managing model latency on MCUs~\cite{banbury2021micronets}. The structure of each block consists of 1$\times$1 Convolutions, 3$\times$3 Depthwise Convolutions, and 1$\times$1 Convolutions. We design each block using a 2D convolutional layer to to effectively handle various input types and extract the initial features. Subsequently, we use a consistent padding strategy to control the depth of OPS, ensuring that the output of each block matches its input. Lastly, we incorporate a linear classifier in each block as the output layer for single-event detection.

To define the model search space for efficient architectures on edge devices, we configure channel sizes \( L \) (ranging from 32 to 512) and OPS sizes \( O \) (3 to 7), drawing from models like MobileNet~\cite{sandler2018mobilenetv2}, DSCNN~\cite{zhang2017hello} for mobile devices, and MicroNets~\cite{banbury2021micronets} for MCUs. This leads to 60 potential configurations (\( N = L \times O \)), each comprising three sub-networks. Our objective is to identify the optimal configuration \( N^* \) that balances minimal OPS with maximal accuracy. As outlined in Algorithm~\ref{alg:search} (Lines 1-9), the search process involves initially setting a best backbone and score (Line 1), iterating through combinations of channel and OPS sizes (Lines 2-3), and assessing candidate NNs based on accuracy and operational space trade-offs (Lines 5-8), to ultimately select the most efficient and accurate NN backbone (Line 9).

% \vspace{-1em}
\subsection{Multiple-event Sharing}
\label{subsec:architecturesearch}

For a $C$ multi-event detection task, a common approach is to develop individual models to ensure reusability across different applications or use cases and to optimize efficiency for each model~\cite{pham2022pros}. These models can occupy $C$ times the MCU memory and computation cost compared to a single-event model. However, some singular events may share similar characteristics, which can be captured by an identical network for feature extraction. For example, EEG signals are often used to detect alpha waves (event 1) and beta waves (event 2) using two independent models, despite the fact that both waves describe brain activities and can share certain information.

\textbf{\textit{{Using Heads to Share Entire Backbone.}}} We propose our multi-event detection models, which share three sub-networks (i.e., shallow, medium, and deep) and consist of $C*3$ adaptive classifiers (cf. Figure \ref{fig:systemoverview_orig}). Compared to multi-class classification, our multi-event sharing framework allows for more flexibility in single-event detection, which is especially preferred on low-power MCUs to ensure efficiency and reusability across multiple applications. Specifically, as illustrated in Figure~\ref{fig:systemoverview_orig}, for each shared shallow, medium, and deep backbone NNs, we design $C$ independent classifiers to distinguish the $C$ events. Each classifier is composed of an adaptive pooling layer and a linear layer. The adaptive pooling layer aims to adjust the different output sizes from the searched shallow, medium, and deep sub-networks to match the input size of the classifiers. We optimize all the classifiers in a multi-task learning paradigm.

\textbf{\textit{Uncertainty-aware Cascade Learning}.}
% \lo{what about call it Cascade without the final D, as in the original work by Marquez? It sounds also a bit better.}\hj{Fixed. Thanks! I will fix other areas as well including Figure 2} 
To train the aforementioned shallow, medium, and deep models for MCUs, we propose an uncertainty-aware cascade model inspired by deep cascade learning for training our early-exit models. As illustrated in Algorithm~\ref{alg:search} (Lines 10-14), we employ three optimizers for the three exits, with each exit representing one-third of the model layers. Initially, we train the first one-third of the layers in the searched backbone model and then utilize its output to train the second exit. Finally, we optimize the third exit.

\begin{figure*}[!t]
  \centering
  \subfloat[Uncertainty operators using TFLM]{
    \includegraphics[width=2.5in]{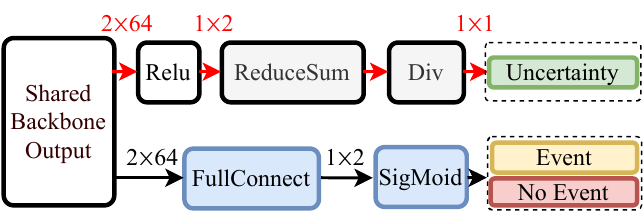}\label{fig:operator_ur2m}}
  % \hfil
  \subfloat[MCUs library optimization]{
    \includegraphics[width=4.5in]{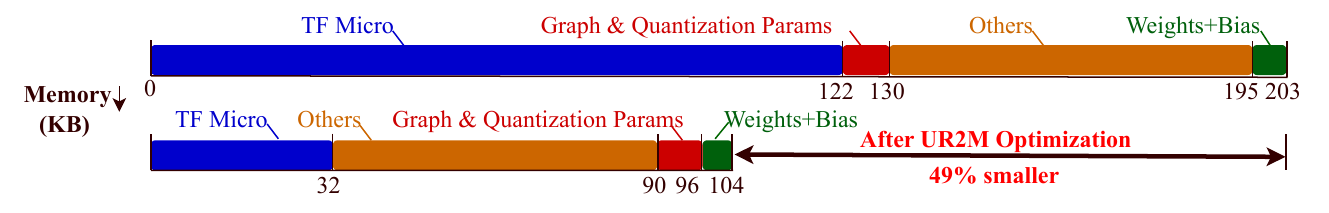}\label{fig:memory_sram}}
  
  \caption{Deployment stage. (a) Uncertainty deployment on MCU based on multiple operators to calculate uncertainty and classification results. (b) MCU library space before optimization (top) and after optimization (bottom).
  }
  \label{fig:implementation}
  \vspace{-1.5em}
\end{figure*}

During each exit, we apply a single-layer linear layer (referred to as a head) for each event, which takes input maps of the output dimensions of the early-exits. Each early exit produces two outputs: the prediction and the uncertainty. We optimize all sub-networks concurrently on the server.

Our design is supported by the MCU libraries of Tensorflow Lite Micro (TFLM) in terms of multi-tenancy (e.g., enabling model deployment in a cascade manner) and memory planner (e.g., reusing the same operator's memory). This coherence can significantly reduce the overheads compared to the conventional multi-event detection models. Overall, our approach aims to optimize the performance of the models while accounting for uncertainty and providing early exits for faster inference.

\section{Implementation}
\label{sec:implementation}
\subsection{System Implementation}
\label{sec:Implementation}

\textbf{\textit{Hardware}.}
The training stage of our system is implemented and tested on a Linux server equipped with an Intel Xeon Gold 5218 CPU and NVIDIA Quadro RTX 8000 GPU. The shared backbone and multiple heads are pre-trained during this stage. Afterwards, in the deployment stage, we deploy the shared backbone and heads on two MCUs. The first one is the STM32F446ZE (or F446ZE), which has an ARM Cortex M4 processor with 128 KB of SRAM and 512 KB of eFlash. The other one is the STM32H747XI (or 747XI), featuring a dual-core processor (ARM Cortex M4 and M7) with 1 MB of SRAM and 2 MB of eFlash. Our evaluation only utilizes one core (ARM Cortex M7) since MCUs are typically equipped with only one CPU core. This setup limits the usage space of SRAM and eFlash to 512 KB and 1 MB, respectively.

We developed and assessed our system's training stage using PyTorch 1.8, and tested various baselines on a Linux server. The evidential uncertainty module is implemented with Python and NumPy. We adopted TensorFlow Lite Micro (TFLM)~\cite{robert_tflm_mlsys2021} for MCU deployment due to its portability, ease of use, and support for numerous neural network layers and optimized kernels. \systemName's deployment stage and online optimization scheme are developed in C++ on two MCUs (ARM Cortex M4 and M7). To deploy a PyTorch model on MCUs, we convert it to TensorFlow Lite (TF Lite) using ONNX representation and the TF Lite converter. The model is run on MCUs using TFLM and Mbed OS. Additionally, the CMSIS-DSP software library processes raw signals to generate model inputs (e.g., MFCC features), and the CMSIS-NN kernels in TFLM facilitate efficient neural network operations on MCUs.

\textbf{\textit{Multi-tenancy Deployment}.} 
To facilitate multi-event sharing on MCUs with limited memory, we develop a multi-tenancy deployment for early-exit models using TFLM. \systemName{} utilizes multiple model interpreters to allocate memory from a unified space, ensuring efficient model operation. During evaluation, this deployment strategy is applied to all baselines and the \systemName{} model. For example, Deep Ensembles have five models, potentially using 5$\times$ eFlash space. However, with optimization, it only consumes 2$\times$ more SRAM (cf. $\S$\ref{subsec:systeem efficiency}) due to multi-tenancy deployment.

\vspace{-0.5em}
\subsection{Uncertainty Operator Implementation}
To capture the uncertainty at inference time on MCUs, we only use TFLM-supported operations\interfootnotelinepenalty=1000. First, we utilize a ReLU operator to regulate the distribution of the output as non-negatives. Then, based on these outputs, we follow Eq.\ref{beta_uncertainty} to generate uncertainties. Specifically, calculating uncertainty first requires the sum of reduced dimensions. Although the {\em reduced\_sum} operator is supported, it is not available for TFLM. To solve this, we use a {\em squeeze} operator to reduce the output dimensions, followed by a {\em sum} operator. Finally, we apply a {\em divide} operator to generate the uncertainty. We wrap the above-mentioned operators within the model and implement them in the TFLM library to save the overhead of uncertainty prediction. The overall uncertainty implementation is shown in Figure~\ref{fig:operator_ur2m}.

\subsection{MCU Library Optimization}\label{subsec:libraryoptimization}

Unlike mobile devices' memory architecture that employs large off-chip main memory (e.g., DRAM), MCUs consist of only small-sized on-chip memory (e.g., SRAM and eFlash) (cf. Figure~\ref{fig:mcu}). To understand the memory requirements of our model to fit in MCUs, we first compute the memory usage of \systemName{}. For a searched shallow model with 8-bit int quantization, we observe that TFLM requires 79 KB of SRAM and 203 KB of eFlash, which falls within the tight memory budgets of many MCUs, for example, 64 KB of SRAM and 128 KB of eFlash of STM32F205VB as described in Figure~\ref{fig:mcu}. In particular, on SRAM, the memory usage includes intermediate tensors (30 KB), persistent buffers (3 KB), runtime overhead of the TFLM interpreter (6 KB), and MBed OS and other libraries (10 KB). Additionally, Figure~\ref{fig:memory_sram} top shows the on-chip eFlash architecture of an F446ZE MCU and how TFLM allocates memory space to run a shallow model on an MCU.

\textit{Note that since we only conduct 8-bit post-quantization, we only observe a maximum of 1\% performance drop between the pre-and post-quantization stages among all methods.}

Given the limited memory space for searching the optimal model parameters, we propose optimizing the TFLM library. First, we removed all operation-related files that did not impact our backbone. Then, we reordered the operations files based on our backbone structure. As shown at the bottom of Figure~\ref{fig:memory_sram}, our MCU library significantly optimized the TFLM interpreter's runtime overhead, reducing it from 122 KB to 32 KB (3.8$\times$ smaller). Moreover, the graph definition was reduced by 2 KB, from 8 KB to 6 KB, in the eFlash memory. After the optimization, a total of 104 KB of memory is used, which can now fit into the STM32F205VB and many other MCUs. Overall, \systemName{} optimizes 49\% of eFlash memory compared to the baseline TFLM library.

\textit{Note that during the evaluation, we applied the same MCU library optimization strategy to all baselines as well as the \systemName{} model.}
\setlength{\tabcolsep}{1pt}
\section{Evaluation settings}
\label{sec:evaluation}

\begin{figure*}[t]
  \centering
  \subfloat[Model sizes \textit{vs.} Acc.]{
    \includegraphics[trim={0.8cm -0.17cm 0.8cm 0.5cm},clip,width=1.35in]{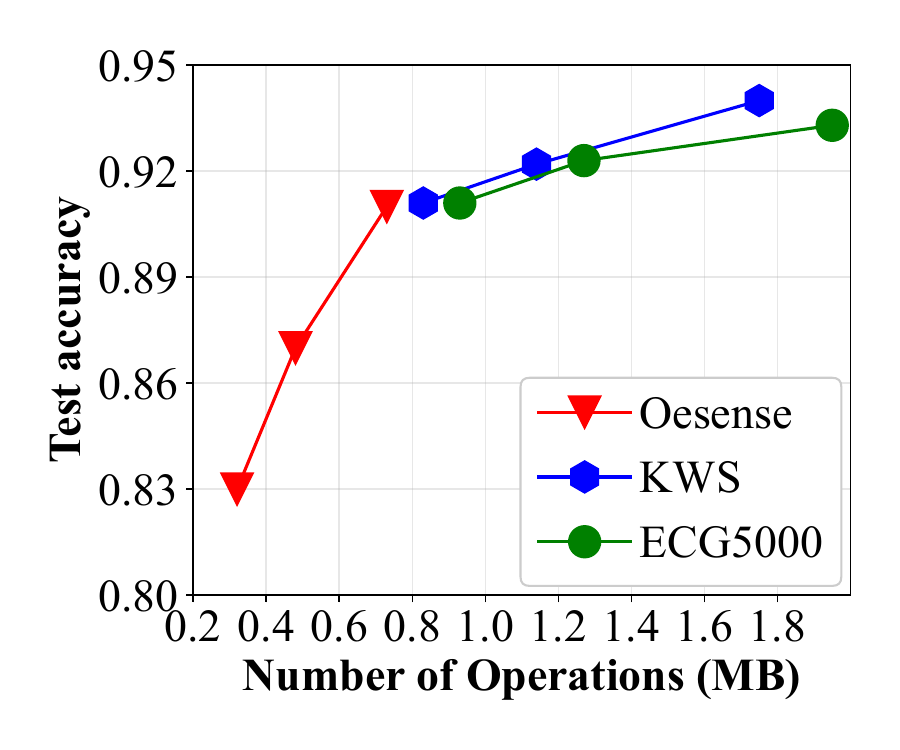}\label{fig:architecture_all}}
  \subfloat[Acc. on Oesense]{
    \includegraphics[trim={0.5cm 0.2cm 0.7cm -0.2cm},clip,width=1.35in]{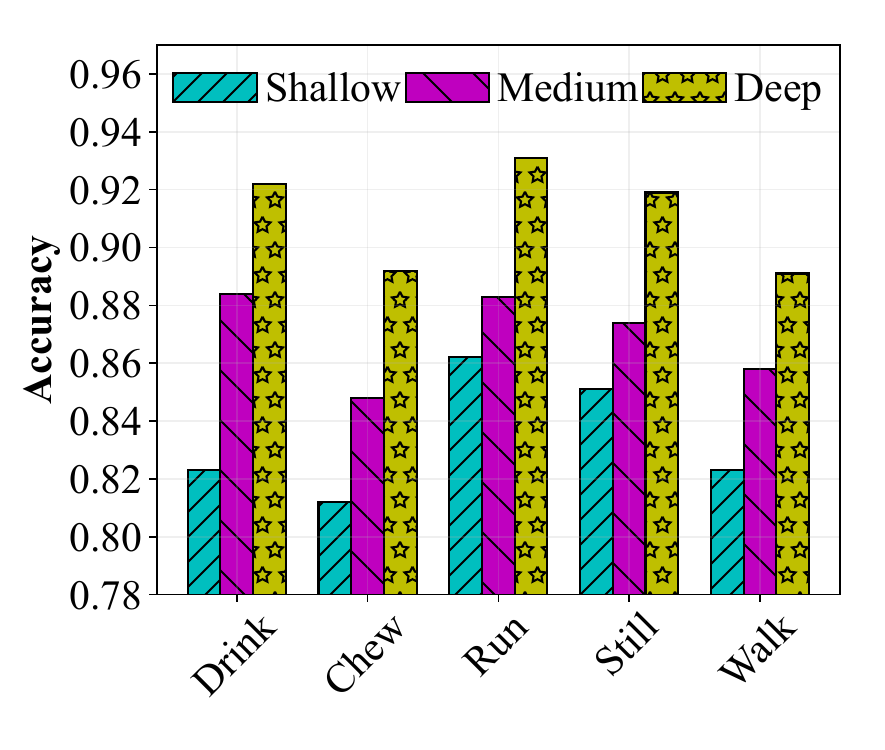}\label{fig:threshold_oesense}}
  \subfloat[Acc. on KWS]{
    \includegraphics[trim={0.5cm 1cm 0.7cm -0.8cm},clip,width=2.55in]{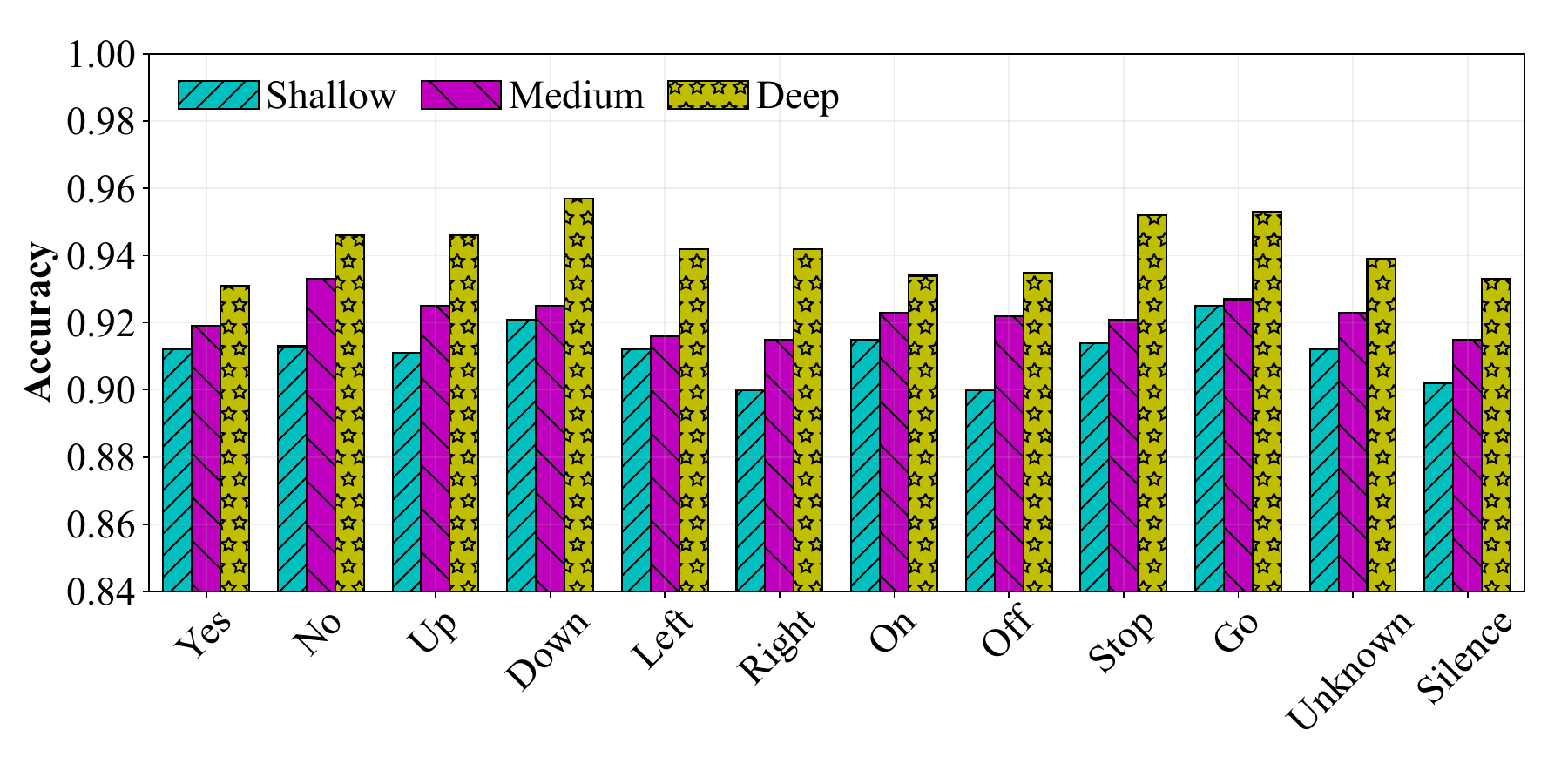}\label{fig:threshold_KWS}}
  \subfloat[Acc. on ECG5000]{
    \includegraphics[trim={0.5cm 0.7cm 0.7cm 0.6cm},clip,width=1.45in]{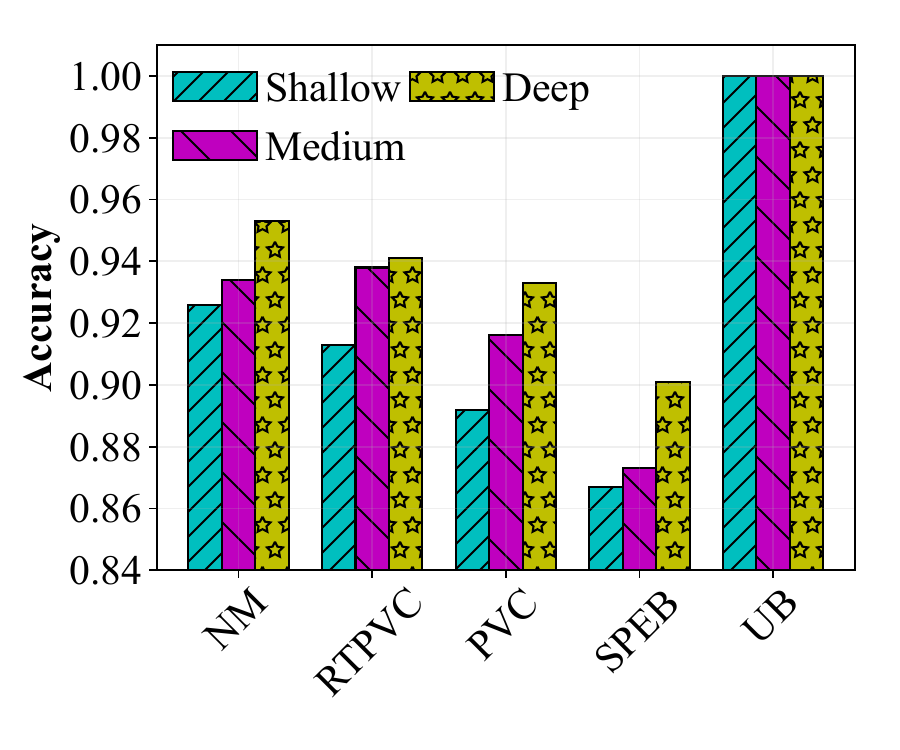}\label{fig:threshold_ecg5000}}
  \caption{Model sizes \textit{vs.} accuracy and early exit result for single events. Note that the ECG5000 UB event has only one test sample.}
  \label{fig:thresholdall}
  \vspace{-1em}
\end{figure*}

\subsection{Evaluated Datasets}\label{subsec:experimental setup}

Our target application scenarios are focused on WED applications. Specifically, we evaluate three wearable datasets, including in-ear activity recognition~\cite{ma2021oesense}, audio event keyword spotting~\cite{warden2018speech}, and heart disorder event detection~\cite{chen2015general}. We experiment with these three datasets, each featuring different data modalities that suit \systemName\ settings. For imbalanced datasets, we use SMOTE~\cite{chawla2002smote} to upsample the training data.

\textbf{\textit{In-ear Dataset}.} Oesense~\cite{ma2021oesense} contains an in-ear audio dataset for activity recognition (including five events: ``walk'', ``run'', ``still'', ``drink'', and ``chew'') among 31 subjects. For preprocessing, we first segment the original audio into one-second segments and set the sampling rate at 4 kHz. Then, we extract the 2-D MFCC features for each segment. 10 MFCC features are then obtained from an audio frame with a length of 80 ms and a stride of 40 ms, yielding an input dimension of 1$\times$10$\times$21. After preprocessing each event, we obtained 40,064 training samples (90\%) and 4,452 test samples (10\%) for all five activities.

\textbf{\textit{KWS Dataset}.} The Keywords Spotting (KWS) V2~\cite{warden2018speech} dataset contains 105,829 utterances from 2,618 speakers. There are 35 words split into 12 classes, including ten keyword spotting classes and an 'unknown' class (remaining 24 words). For preprocessing, we first constrained all event samples to one second by segmentation or zero-padding and set the sampling rate at 16 kHz. Then we extracted MFCC features using 640 FFT points and 320 points of hop length. We obtained 10 MFCC features from an audio frame with a length of 40ms and a sliding window of 20ms, yielding the input dimension of 1$\times$10$\times$51. After preprocessing, we obtained 92,502 total event training samples (90\%) and 10,278 test samples (10\%).

\textbf{\textit{ECG5000 Dataset}.} The ECG5000 dataset~\cite{chen2015general} is a 20-hour long one-channel ECG dataset that contains 92,584 heartbeats, including five different types of heart events: Normal (NM) (58.4\%), R-on-T Premature Ventricular Contraction (RTPVC) (35.3\%), Premature Ventricular Contraction (PVC) (3.9\%), Supra-ventricular Premature or Ectopic Beat (SPEB) (2\%), and Unclassified Beat (UB) (0.5\%). For preprocessing, we resample the input duration of 0.56s with 140 samples into 560 samples. Then we reshape the input into 10 channels, yielding the input dimension of 1$\times$10$\times$56. After the preprocessing, we obtained 4,500 total event training samples (90\%) and 500 (10\%) test samples. Note that UB has only one test sample.

\subsection{Uncertainty Metrics}
We compare \systemName{} using three important uncertainty metrics: Brier score, Negative Log-Likelihood (NLL), and Expected Calibration Error (ECE), to examine the uncertainty estimation performance.

\subsection{Uncertainty Quantification Baselines}
\label{baselines}
We evaluate the proposed method by comparing it to three baseline uncertainty solutions: the traditional softmax-based models, deep ensembles and data augmentation. It is important to note that MCDP~\cite{gal2016dropout} is not available for the MCUs library TFLM because it stores models as binary files that cannot be modified. Moreover, its computational costs are similar to or greater than deep ensembles, while its uncertainty performance is lower than that of deep ensembles~\cite{qendro2021benefit}.

\textbf{\textit{Vanilla EDL}}. Vanilla EDL~\cite{sensoy2018evidential} is the state-of-the-art (SOTA) model to \textit{efficiently} quantify uncertainty and can be implemented on MCUs.

\textbf{\textit{Deep Ensembles}}. Deep ensembles approach (denoted as D(Softmax)+Ense)~\cite{lakshminarayanan2017simple} is the SOTA model to \textit{accurately} quantify uncertainty estimation, which typically ensembles $N$ deterministic Softmax models with random weight initializations. We use $N=5$ which is widely adopted in recent efficient studies~\cite{qendro-early-2021}.

\textbf{\textit{Data Augmentation}}. Test time data augmentation (denoted as D(Softmax)+InAug)~\cite{wen2020time} is a \textit{memory-efficient} uncertainty quantification method generating multiple test samples by applying data augmentation techniques through a single model. We utilize five augmented samples, incorporating Jittering, with a mean $\varepsilon$ of 0 and a standard deviation $\sigma$ of 0.03, which are added to the test data. 
% \vspace{-1em}
\section{Results}
\label{sec:results}
This section will discuss the results and answer the following questions: (1) How efficient is \systemName\ for typical MCUs? (2) How robust is \systemName\ compared with traditional point prediction models? 

\subsection{Performance of Event Detection}
\label{subsec:model_architecture}

Utilizing the Adam optimizer with a learning rate of $1\mathrm{e}{-3}$, a 32 batch size, and an early stopping of 5 epochs, we train our networks, showcased in Figure~\ref{fig:architecture_all} and Figures~\ref{fig:threshold_oesense}-~\ref{fig:threshold_ecg5000}. While system accuracy generally increases with OPS across all datasets, significant increases in overhead do not invariably equate to notable accuracy improvements, as observed in the KWS and ECG5000 datasets. For instance, a shallow Oesense model (accuracy: 0.83, parameters: 0.38 MB) contrasts with the medium and deep models, which respectively present 0.87/0.58 MB and 0.91/0.76 MB in accuracy/parameters. The 2\% accuracy enhancement when transitioning from medium to deep models incurs a 31\% overhead spike. Similarly, for ECG5000, a 1\% accuracy improvement requires doubling the model sizes. Shallow models across all datasets exhibit proficient performance (e.g., $>$80\%) with minimized model size, hinting that \systemName{} could deliver effective performance with modest overheads.

Regarding the channel sizes, our searched model yields the output shape for each OPS as [5,11] for Oesense, [5, 26] for KWS, and [5, 29] for ECG5000, respectively. Figure~\ref{fig:thresholdall} further illustrates the \systemName{} performance for single event detection using shallow, medium and deep network structures.

\begin{figure}[t]
  \centering
  \centering
  \includegraphics[width=1.05\columnwidth]{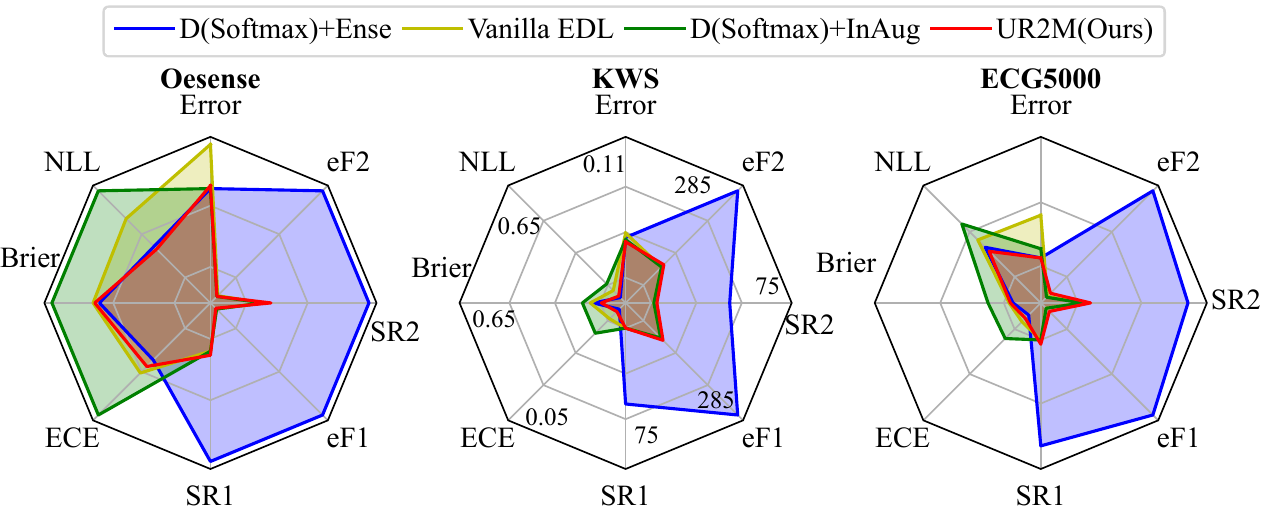}
  \vspace{-2em}
  % \ftodo{\S 8C1 \#R1}
  \caption{Comparing Vanilla EDL, data augmentation, deep ensembles (SOTA), and \systemName{} using uncertainty, error rate, and memory usage metrics across three datasets. eF1 and SR1 refer to the eFlash and SRAM usage of H747XI, while eF2 and SR2 refer to those of F464ZE, respectively. For all metrics, lower values are preferred.}
  \label{fig:radar}
\end{figure}

Based on Figure~\ref{fig:radar}, we can observe that \systemName{}'s uncertainty metrics are better than Data Augmentation (D(Softmax)+InAug) baseline across all three datasets, with up to 22\% lower NLL scores (0.65 to 0.53). This improvement indicates that the proposed method produces better-calibrated models that are less prone to overconfidence errors. Compared to the D(Softmax)+Ense model, \systemName\ achieves similar performance in terms of both uncertainty estimation and prediction accuracy. For instance, \systemName{} outperforms D(Softmax)+Ense by 8.0\% in terms of Brier score for KWS, and achieves 1.7\% and 2.4\% relative improvements in NLL for Oesense and ECG5000, respectively.

\textit{Notably, \systemName\ achieves these results while using up to only half of the memory, much less energy, and latency required by SOTA method deep ensembles (cf. Figure~\ref{fig:radar} and \S\ref{subsec:systeem efficiency}), demonstrating the computational efficiency of \systemName\ without compromising uncertainty estimation.}

\vspace{-0.5em}
\subsection{Impact of different Uncertainty Thresholds}
\label{subsec:effect}

\begin{figure}[t]
    \centering
    \begin{minipage}{0.5\columnwidth}
        \centering
        \includegraphics[trim={0.9cm 0.7cm 0.7cm 0.7cm},clip,height=2.6cm]{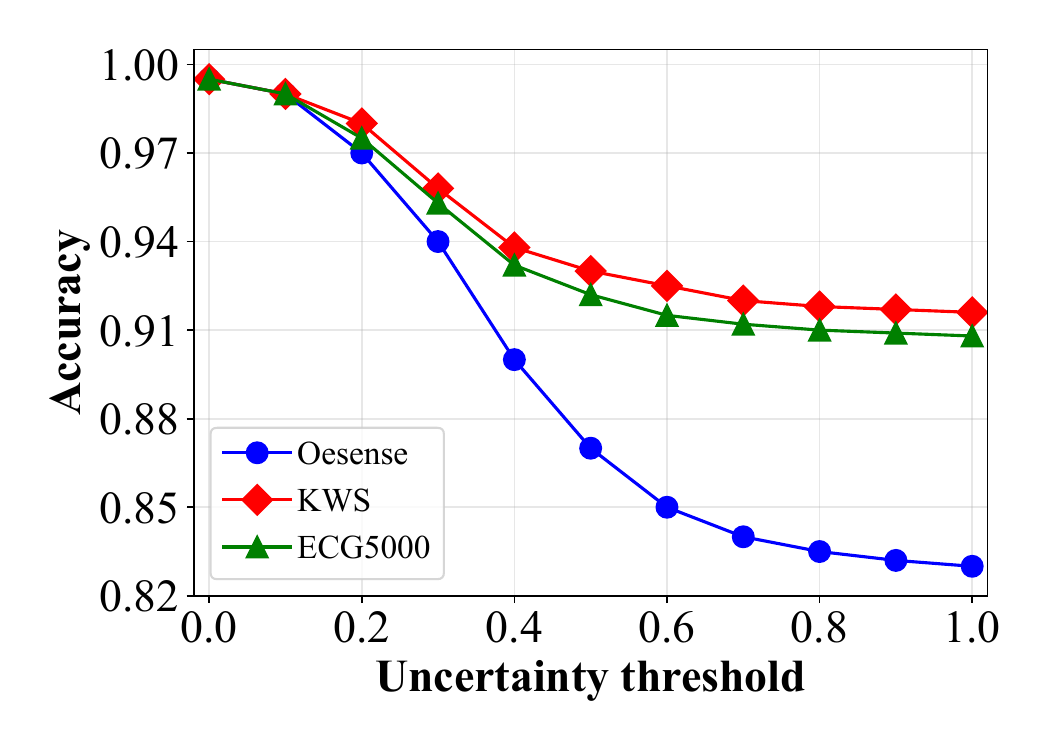}
        \caption{Uncertainty impact}
        \label{fig:threhold_accuracy}
    \end{minipage}
    \hspace{-.7em}
    \begin{minipage}{0.5\columnwidth}
        \centering
        \includegraphics[trim={0cm 0cm 0cm 0cm},clip,width=4cm]{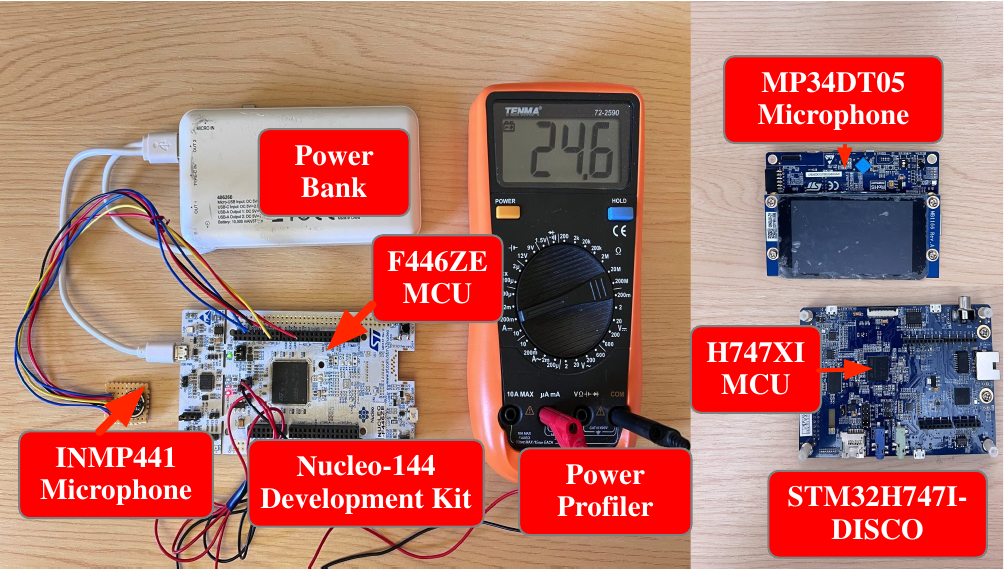}
        \vspace{1em}
        \caption{End-to-end deployment}
        \label{fig:setup}
    \end{minipage}
    
\end{figure}

Users can decide on the uncertainty threshold according to their specific applications. For example, in healthcare applications (e.g., heart attack detection), we prefer a low uncertainty (e.g., \( u \)=0.05) for detected heart attacks to avoid disastrous consequences. This tradeoff is depicted in Figure~\ref{fig:threhold_accuracy}. In other scenarios (e.g., running detection), a higher uncertainty threshold can be tolerated to save battery life by exiting through shallow layers. Similarly, it is observed that increasing the threshold gradually reduces latency across all three datasets when evaluated on the F446ZE and H747XI MCUs. With a higher uncertainty threshold, more samples are filtered out by the shallow and medium sub-networks, and fewer samples pass through deep models, leading to reduced latency. This indicates that selecting different uncertainty thresholds allows users to obtain a personalized model, increasing the usability of \systemName{}.

\textit{In sum, our model design, which allows users to define the threshold, can help determine the optimized threshold to balance the tradeoff, thereby achieving personalized models.}

\subsection{End-to-end System Efficiency}
\label{subsec:systeem efficiency}

Following the optimization of all baselines and \systemName\ using techniques including multi-tenancy deployment, model quantization, and MCU library optimization, we evaluate their runtime efficiency during deployment on MCUs (Figure~\ref{fig:setup}). Our evaluation encompasses the entire system, including signal acquisition, feature extraction, and memory usage in terms of SRAM and eFlash required for model execution. We conducted experiments with various datasets and two typical resource-constrained MCUs, the F446ZE and H747XI. Although the focus is primarily on the ECG5000 dataset due to page limits, note that consistent outcomes were observed across all three datasets.

\textbf{\textit{Model Inference Memory Footprint}.}
Based on our implementation, \systemName\ consumes only 49 KB and 51 KB of SRAM (38.5\% and 9.9\% of the total SRAM of F446ZE and H747XI, respectively) as shown in Figure~\ref{fig:radar}. Additionally, as shown in Figure~\ref{fig:radar}, \systemName\ requires 142 KB and 145 KB of eFlash (27.7\% and 14.1\% of the total eFlash of F446ZE and H747XI, respectively). These results demonstrate that \systemName\ consumes only a small portion of the limited resources of MCUs, leaving enough resources for other applications to be supported simultaneously. Furthermore, \systemName\ requires only 66-67\% of SRAM (49 KB vs. 75 KB for F446ZE and 51 KB vs. 75 KB for H747XI) and 51\% of eFlash (142 KB vs. 280 KB for F446ZE and 145 KB vs. 283 KB for H747XI) compared to the deep ensembles baseline.

\textbf{\textit{Signal Acquisition Overheads.}}\label{appendix:signal_acquisition} To evaluate signal acquisition overheads for the F446ZE MCU, we employ an INMP441 MEMS microphone. For the H747XI, we use the MP34DT05-A built-in microphone on the H747I-DISCO evaluation board (Figure~\ref{fig:setup}). We assess energy consumption and memory usage as key factors. Energy consumption (J) is computed as the product of time/latency (t) and power (W). Power is determined from input voltage (V) and current measurements (A), conducted with a Fluke 87V digital multimeter. For the F446ZE, we record a power consumption of 24.6 mA at 3.3V, resulting in 81.18 mW for one second of audio signal acquisition. Memory-wise, it uses 4KB of SRAM and 32KB of eflash. In contrast, the H747XI consumes 31.6 mA at 3.3V, totaling 104.28 mW in power. It utilizes 29KB of SRAM and 66KB of eflash. Overall, signal acquisition overheads for these two MCUs are minimal.

\textbf{\textit{Feature Extraction Overheads.}}\label{appendix:evaluation} 
The feature extraction step for both \systemName\ and the baselines is the same, using MFCC features as inputs. The extraction process is fast, taking only 4.505 ms and 10.913 ms per extraction for the H747XI across two datasets, indicating minimal overhead.

\begin{figure}[t]
  \centering
  \centering
  \includegraphics[width=1.02\columnwidth]{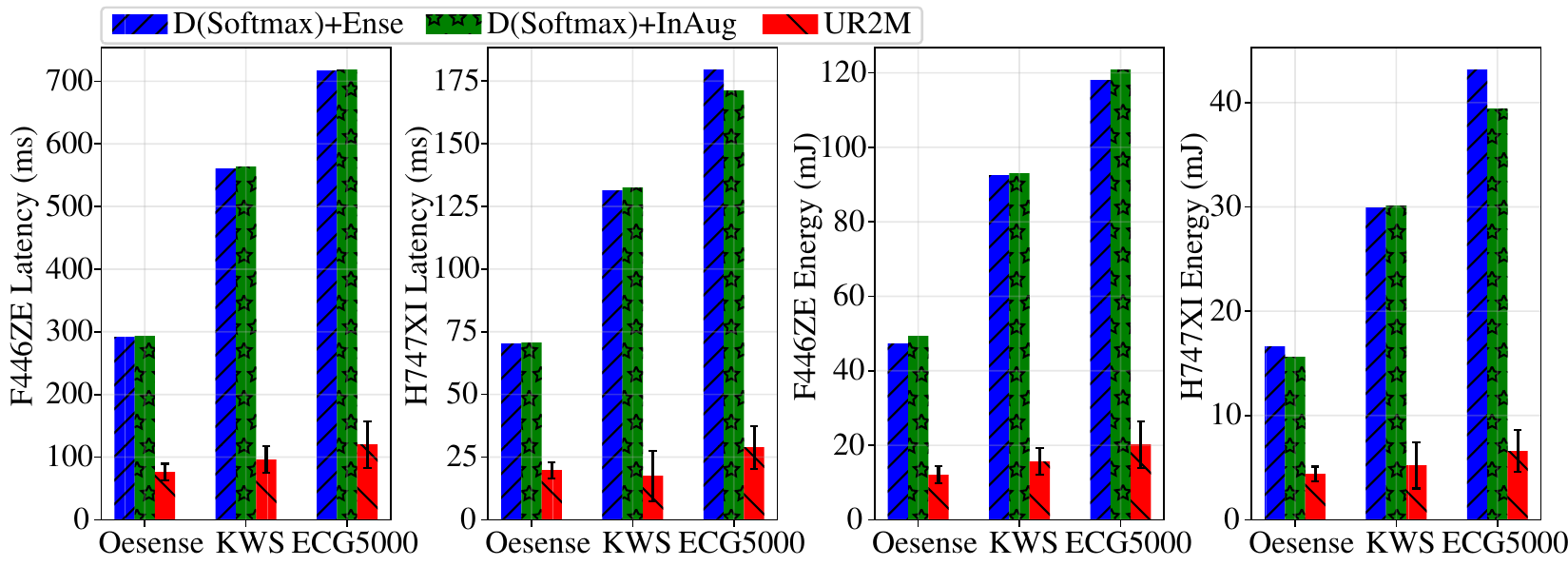}
  \vspace{-1em}
  \caption{Comparison of latency and energy consumption of uncertainty-aware methods on two MCUs.}
  \label{fig:latency}
\end{figure}

\textbf{\textit{Model Inference Latency}.}
Using the MBed Timer API to measure latency on MCUs, Figure~\ref{fig:latency} illustrates \systemName's and baseline inference results across three datasets and two MCUs. With uncertainty thresholds ($u$) ranging from 1 to 0, \systemName presents latencies from lowest to highest, respectively. While baseline approaches, like deep ensemble, yield reliable uncertainty estimations, they exhibit high inference latencies of 717.2-717.4 ms on F446ZE and 171.1-179.3 ms on H747XI per sample. Conversely, \systemName ensures both reliable uncertainty and minimized latency, cutting inference latencies up to 864\% (83.0 ms vs. 717.2 ms) on F446ZE and 835\% (20.2 ms vs. 171.1 ms) on H747XI. Moreover, \systemName enhances latency by approximately 456\% against other baselines, even without uncertainty filtering.

\textbf{\textit{Model Inference Energy Consumption}.}
Similar to the latency results, \systemName\ significantly reduces energy consumption compared to the baselines, as shown in Figure~\ref{fig:latency}. For example, \systemName\ decreases energy consumption by up to 834\% (116.0 mJ vs. 13.9 mJ) on F446ZE and 857\% (39.4 mJ vs. 4.6 mJ) on H747XI when compared to the best-performing benchmark uncertainty-aware baselines. Also, we observe that \systemName\ achieves around 450\% energy improvement compared to the baselines without uncertainty filtering.

\subsection{Robustness Against Signal Uncertainties}
\begin{figure}[t]
  \centering
  \includegraphics[trim={0.5cm 0cm 0cm 0cm},clip,height=0.45\columnwidth]{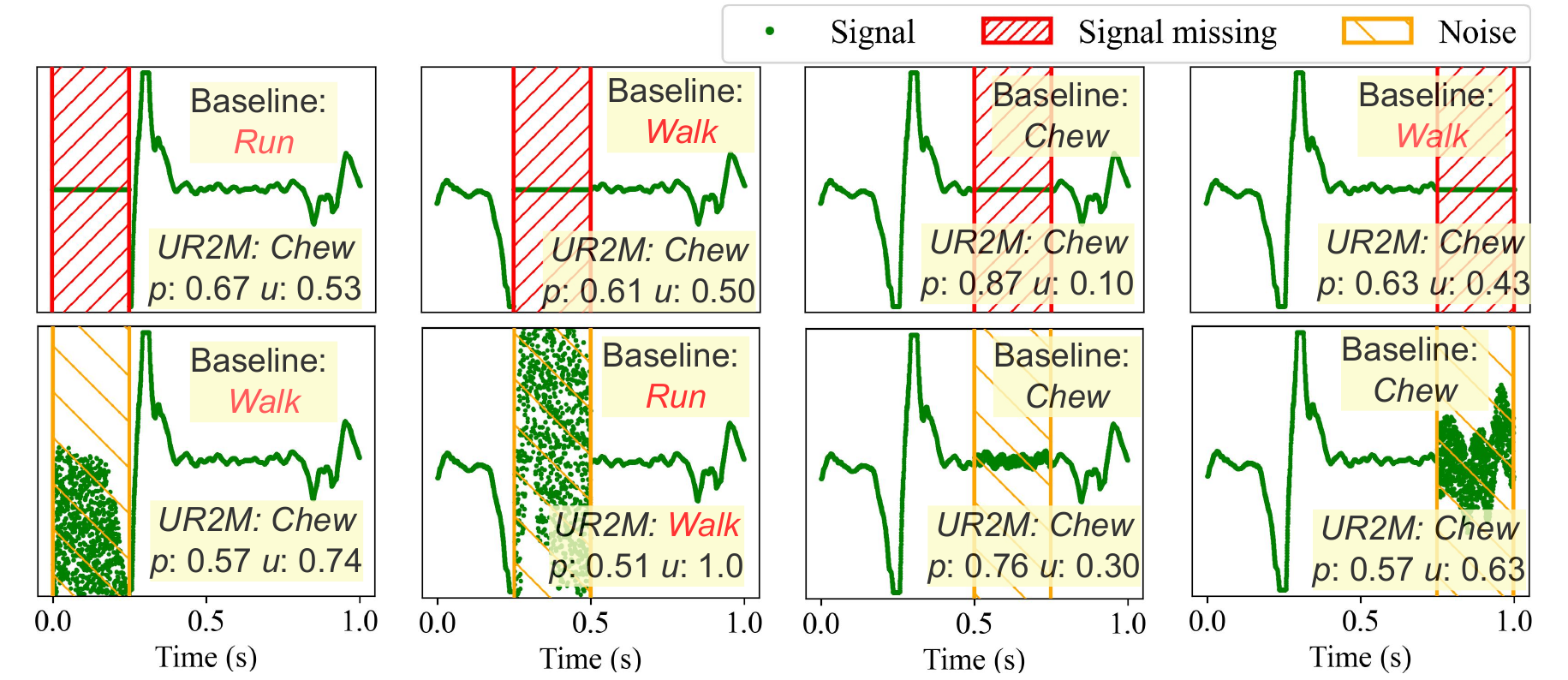}
  \vspace{-1em}
  \caption{Uncertainty estimation towards signal missing and noise. Labels in red indicate wrong predictions.}
  \label{fig:uncertainty_simulation}
\end{figure}

We evaluate \systemName\ in the context of two types of signal uncertainties: signal missing (replace as zero) and noise (gaussian noise). Due to page limitations, we compare our method with traditional softmax-based NNs having the same model structure. As demonstrated in Figure~\ref{fig:uncertainty_simulation}, for a correct event signal ``Chew'', the absence of signal and the presence of random noise can lead softmax-based NNs to predict incorrectly. In contrast, \systemName\ can accurately predict most corrupted signals. When predictions are incorrect, \systemName\ also exhibits high uncertainty (e.g., $u$=1.0), which could be used for alerting the system to potential misclassifications or triggering additional validation steps.

\label{sec:autofeature}
\vspace{-0.5em}
\section{Discussion}
\label{sec:discussion}

In this section, we discuss several possible future directions for our work.

\textbf{Generalizing \systemName\ to other sensors and higher-end MCUs.}
Ideally, \systemName\ could be generalized to any wearable sensors driven by MCUs. However, sensor signal complexity and limited MCU memory size pose limitations. More complex signals usually require larger model sizes, challenging the deployment on the constrained memory of MCUs. Fortunately, recent work~\cite{pham2022pros} shows that by investigating compressive sensing, key patterns of primitives in signals can be compressed and extracted, which indicates it can reduce the model size to save system overhead. Therefore, we will study how compressive sensing combined with \systemName\ could further reduce system overhead to generalize to ultra low-end MCUs. We envision our method could also benefit higher-end MCUs, e.g. STM32F4, which has 1MB flash and 192KB SRAM. Since less memory is required, higher-end MCUs could experience improvements in latency and energy efficiency.

\textbf{Impact of \systemName\ on future WED systems.} Our work has illustrated that uncertainty is a key criterion to ensure reliable prediction in WED systems. Therefore, an important and urgent question is how to define uncertainty tolerance thresholds for specific applications. Fortunately, for healthcare applications, we can design this criterion through a doctor-in-the-loop strategy to select the optimal threshold. 
\section{Conclusion}
\label{sec:conclusion}
In this paper, we have proposed \systemName{}, a resource and uncertainty-aware framework which can efficiently and reliably enable wearable event detection and related uncertainty on MCUs. By exploiting evidential uncertainty theory, cascade learning, and system optimization, \systemName{} significantly improves energy and memory efficiency for MCUs without sacrificing accuracy, enabling real-time and reliable event detection.

\section{Acknowledgment}
This work is supported by ERC through Project 833296 (EAR), and Nokia Bell Labs through a donation.

% % \balance
\bibliography{main}
\bibliographystyle{unsrt}
% \appendix

\end{document}